\documentclass{article}
\pdfoutput=1
\usepackage[final]{nips_2017}
\usepackage{showlabels}

\usepackage[dvipdfmx]{graphicx}
\usepackage{bmpsize}
\usepackage[tight,ugly]{units}
\usepackage[utf8]{inputenc} 
\usepackage{pgfplots} 
\pgfplotsset{compat=newest}
\usepgfplotslibrary{groupplots}
\usepackage[T1]{fontenc}    
\usepackage{hyperref}       
\usepackage{url}            
\usepackage{booktabs}       
\usepackage{amsmath, amssymb, amsfonts}       
\usepackage{nicefrac}       
\usepackage{microtype}      
\usepackage{mathtools}
\usepackage{dsfont}
\usepackage{color}
\usepackage{appendix}
\usepackage{caption}
\usepackage{varwidth}

\usetikzlibrary{external}
\usetikzlibrary{spy}
\usepackage[colorinlistoftodos]{todonotes}

\newcommand{\figpath}{experiments/figs}%
 \newlength\figureheight
 \newlength\figurewidth

 \usepackage{array}

\renewcommand{\vec}[1]{{\boldsymbol{{#1}}}} 
\newcommand{\mat}[1]{{\boldsymbol{{#1}}}} 
\newcommand{\R}[0]{\mathds{R}} 
\newcommand{\inv}{^{-1}} 

\ifxetex

\else

\fi

\newcommand{\given}{\,|\,}

\title{Identification of {G}aussian Process State Space Models}

\author{
	\!\!Stefanos Eleftheriadis$^\dagger$,~ Thomas F.W.~Nicholson$^\dagger$,~ Marc P.~Deisenroth$^{\dagger\ddagger}$,~ James Hensman$^\dagger$\\
  $^\dagger$PROWLER.io, \quad $^\ddagger$Imperial College London\\
  \{stefanos, tom, marc, james\}@prowler.io\\
}

\begin{document}

\maketitle

\begin{abstract}
The Gaussian process state space model (GPSSM) is a
non-linear dynamical system, where unknown transition and/or
measurement mappings are described by GPs. Most research in GPSSMs has
focussed on the state estimation problem, i.e., computing a posterior of the latent state given
the model. However, the key challenge
in GPSSMs has not been satisfactorily addressed yet: system identification, i.e., learning the model.
To address this challenge, we impose a structured Gaussian 
variational posterior distribution over the latent states, which
is parameterised by a recognition model in the form of a bi-directional
recurrent neural network. Inference with this structure allows us to
recover a posterior smoothed over sequences of data. We
provide a practical algorithm for efficiently computing a lower bound
on the marginal likelihood using the reparameterisation trick.  This
further allows for the use of arbitrary kernels within the GPSSM. We
demonstrate that the learnt GPSSM can efficiently generate plausible future
trajectories of the identified system after
only observing a small number of episodes from the true system.
\end{abstract}

\section{Introduction}

State space models can effectively address the problem of learning
patterns and predicting behaviour in sequential data. Due to their
modelling power they have a vast applicability in various domains of
science and engineering, such as robotics, finance,
neuroscience, etc.~\citep{brown1998statistical}.

Most research and applications have focussed on linear state space
models for which solutions for inference (state estimation) and
learning (system identification) are well established~\citep{Kalman1960,
  Ljung1999}. In this work, we are interested in non-linear 
state space models. In particular, we consider the case where a
Gaussian process (GP)~\citep{Rasmussen2006} is responsible for modelling the underlying
dynamics. This is widely known as the Gaussian process state
space model (GPSSM).
We choose to build upon GPs for a number of reasons. First,
they are non-parametric, which makes them effective in learning
from small datasets. This can be advantageous over well-known parametric models
(e.g., recurrent neural networks---RNNs), especially in situation where data
are not abundant. Second, we want to take advantage of the probabilistic
properties of GPs. By using a GP for the latent transitions,
we can get away with an approximate model and learn a distribution over
functions. This allows us to account for model errors
whilst quantifying uncertainty,
as discussed and empirically shown by \cite{Schneider1997} and \cite{ Deisenroth2015}.
Consequently, the system will not become overconfident in regions of the space
where data are scarce.

System identification with the GPSSM is a challenging task.
This is due to un-identifiability issues:
both states and transition functions are unknown. Most work so far has focused only on 
state estimation of the GPSSM. In this paper, we focus on addressing the challenge
of system identification and based on recent work by
\citet{frigola2014variational} we propose a novel inference method
for learning the GPSSM. We approximate
the entire process of the state transition function by employing the
framework of variational inference. We assume a
Markov-structured Gaussian posterior distribution over the latent
states. The variational posterior can be naturally combined with a
recognition model based on bi-directional recurrent neural networks,
which facilitate smoothing of the state posterior over the data sequences.
We present an efficient algorithm based on the
reparameterisation trick for computing the lower bound on the marginal
likelihood. This significantly accelerates learning of the model and
allows for arbitrary kernel functions.

\section{Gaussian process state space models}
We consider the dynamical system
\begin{equation}
  \vec{x}_t = f(\vec x_{t-1}, \vec a_{t-1}) + \vec{\epsilon}_f, \quad
  \vec{y}_t = g({\vec{x}}_t) + \vec{\epsilon}_{g},
              \label{eq:ssm}
\end{equation} 
where $t$ indexes time, $\vec x\in\R^D$ is a latent state, $\vec a\in\R^P$ are control
signals (actions) and $\vec y\in\R^O$ are
measurements/observations. We assume i.i.d. Gaussian system/measurement
noise
$ \vec{\epsilon}_{(\cdot)} \sim
\mathcal{N}\big(\vec{0},\sigma^2_{(\cdot)}\vec{I}\big)$. The
state-space model in eq.~\eqref{eq:ssm} can be fully described by the
measurement and transition functions, $g$ and $f$.

The key idea of a GPSSM is to model the transition function $f$
and/or the measurement function $g$ in eq.~\eqref{eq:ssm} using GPs,
which are distributions over functions. A GP is fully specified by a mean $\eta(\cdot)$
and a covariance/kernel function $k(\cdot,\cdot)$, see e.g.,~\citep{Rasmussen2006}. 
The covariance function allows us to encode basic structural assumptions of the class of functions we want to model, e.g., smoothness, periodicity or stationarity. 
A common choice for a covariance function is the radial basis function (RBF).

Let $f(\cdot)$ denote a GP random function, and $\vec X =
[\vec x_i]_{i=1}^N$ be a series of points in the domain of that
function. Then, any finite subset of function evaluations, $\vec f = [f(\vec x_i)]_{i=1}^N$, are jointly Gaussian distributed
%
\begin{equation}
	p(\vec f|\mat X) = \mathcal N \big(\vec f \given \vec \eta,\,\vec K_{xx}\big)\,,
\end{equation}
where the matrix $\vec K_{xx}$ contains evaluations of the kernel
function at all pairs of datapoints in $\mathbf X$, and
$\vec \eta = [\eta(\vec x_i)]_{i=1}^N$
is the prior mean function.
This property leads to the widely used GP regression
model: if Gaussian noise is assumed, the marginal
likelihood can be computed in closed form, enabling
learning of the kernel parameters.
By definition, the conditional distribution of a GP is another GP. If we are to observe the values $\vec f$ at the input locations $\vec X$, then we predict the values elsewhere on the GP using the conditional
\begin{equation}
	f(\cdot) \given \vec f \sim \mathcal {GP}\big(
	\eta(\cdot) + k(\cdot,\mat X)\mat K_{xx}\inv (\vec f - \vec\eta)),\,
	k(\cdot, \cdot) - k(\cdot,\mat X)\mat K_{xx}\inv k(\mat X, \cdot)\big)\,.
\end{equation}
Unlike the supervised setting, in the GPSSM, we are presented with neither values of the function on which to condition, nor on {\em inputs} to the function since the hidden states $\vec x_t$ are latent. The challenge of inference in the GPSSM lies in dually inferring the latent variables $\vec x$ and in fitting the Gaussian process dynamics $f(\cdot)$.

In the GPSSM, we place independent GP priors on the transition function $f$ in eq.~\eqref{eq:ssm} for each output dimension of $\vec x_{t+1}$, and collect realisations of those functions in
the random variables $\vec{f}$, such that
\begin{align}
	f_d(\cdot) \sim \mathcal{GP}\big(\eta_d(\cdot),\,k_d(\cdot,\cdot)\big), \quad \vec{f}_t = [f_d(\tilde{\vec{x}}_{t-1})]_{d=1}^D \quad \text{and}\quad 
	p(\vec{x}_t | \vec{f}_t) = \mathcal{N}(\vec{x}_t | \vec{f}_t, \sigma^2_f\vec{I}),\label{eq:prior}
\end{align}
where we used the short-hand notation $\tilde{\vec x}_t= [\vec x_t, \vec
a_t]$ to collect the state-action pair at time $t$.
In this work, we use a mean function that keeps the state constant, so $\eta_d(\tilde{\vec x}_t) = \vec x_t^{(d)}$.


To reduce some of the un-identifiability problems of GPSSMs, we assume
a linear measurement mapping $g$ so that the data conditional is
\begin{equation}
	p(\vec{y}_t|\vec{x}_t) = \mathcal{N}(\vec{y}_t|\vec{W}_{\!\!g}\vec{x}_t + \vec{b}_{g}, \sigma^2_g\vec{I})\,.
\end{equation}
The linear observation model $g(\vec x) = \vec W_{\!\!g}\vec x + \vec b_g + \vec\epsilon_g$ is not limiting since a non-linear $g$ could be replaced by additional dimensions in the state space \citep{frigola2015bayesian}.

%
%

\subsection{Related work}
State estimation in GPSSMs has been proposed by \cite{Ko2009} and \cite{
Deisenroth2009a} for filtering and by~\cite{Deisenroth2012} and \cite{Deisenroth2012d} for smoothing
using both deterministic (e.g., linearisation) and stochastic (e.g., particles) approximations.
These approaches focused only on inference in learnt GPSSMs and not on system identification, 
since learning of the state transition function $f$ without observing the
system's true state $\vec{x}$ is challenging. 


Towards this approach, \cite{Wang2008}, \cite{Ko2009a} and \cite{Turner2010} proposed methods for learning GPSSMs
based on maximum likelihood estimation. \citet{frigola2013bayesian} followed a Bayesian
treatment to the problem and proposed an inference mechanism based on particle Markov chain
Monte Carlo. Specifically, they first obtain sample trajectories from the smoothing distribution 
that could be used to define a predictive density via Monte Carlo integration. Then, conditioned
on this trajectory they sample the model's hyper-parameters. This approach 
scales proportionally to the length of the time
series and the number of the particles.
To tackle this inefficiency, \citet{frigola2014variational} suggested a hybrid inference approach
combining variational inference and sequential Monte Carlo. Using the sparse variational 
framework from \citep{titsias2009variational} to approximate the GP led to a tractable 
distribution over the state transition function that is  independent of the length of the time series.

An alternative to learning a state-space model is to follow an
autoregressive strategy
\citep[as in][]{murray2001gaussian,likar2007predictive, Turner2011,roberts2013gaussian,
kocijan2016modelling}, to directly model the mapping from previous to current observations. 
This can be problematic since noise is propagated through the system during inference.
To alleviate this, \cite{mattos2015recurrent} proposed the recurrent GP, a non-linear dynamical 
model that resembles a deep GP mapping from observed inputs to observed outputs, with an
autoregressive structure on the intermediate latent states. They further followed the idea 
by \cite{dai2015variational} and introduced an RNN-based recognition model to approximate 
the true posterior of the latent state. 
A downside is the requirement to feed future actions forward into the RNN during inference,
in order to propagate uncertainty towards the outputs.
Another issue stems from the model's inefficiency in analytically computing expectations of the 
kernel functions under the approximate posterior when dealing with high-dimensional latent states.
Recently, \cite{al2016learning}, introduced a recurrent structure to the manifold GP
\citep{Calandra2016}. They proposed to use an LSTM in order to map the observed inputs onto a 
non-linear manifold, where the GP actually operates on. For inefficiency, they followed an approximate inference scheme based on Kronecker products over Toeplitz-structured kernels.

\section{Inference}
\begin{figure}[t]
\begin{tabular}{cc}
\centering
	\includegraphics[scale=.22]{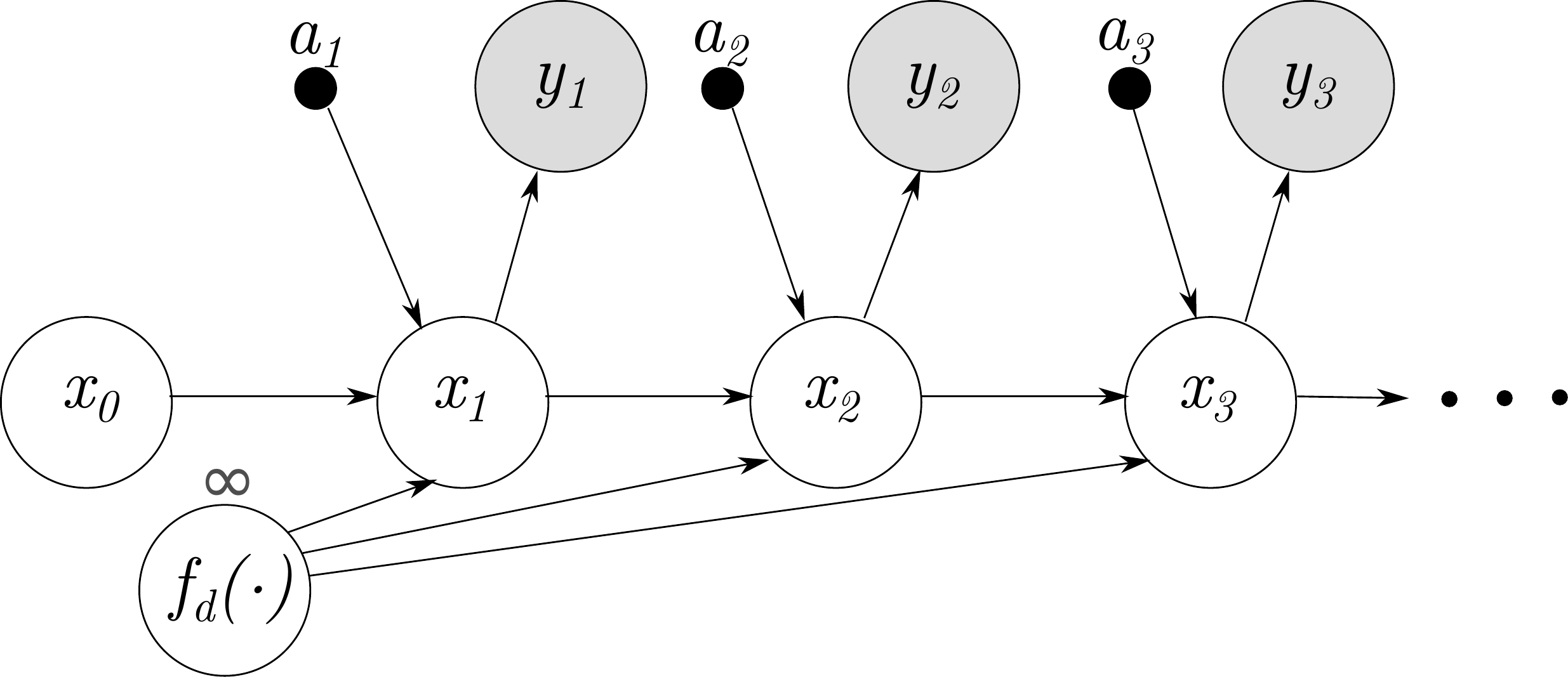} & \hspace*{1cm}\includegraphics[scale=.22]{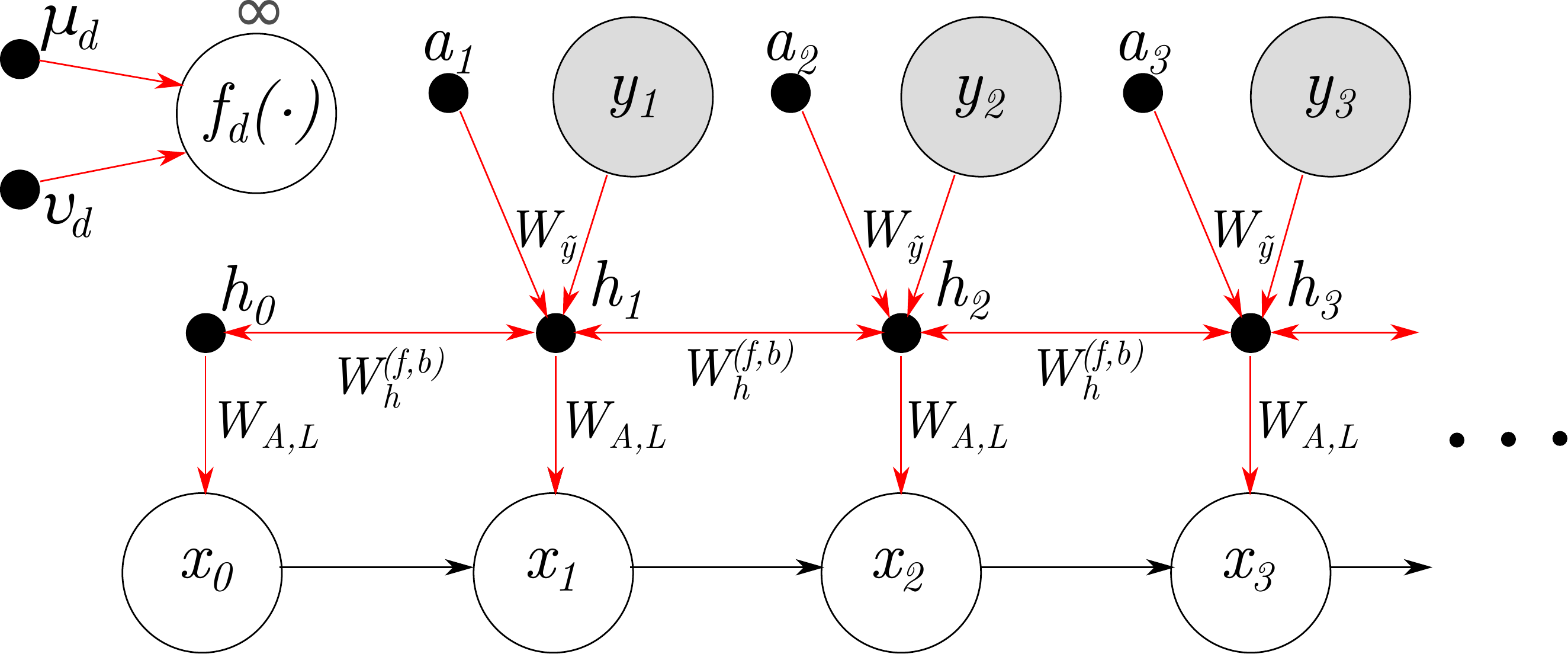}
\end{tabular}
\caption{{\small The GPSSM with the GP state transition functions (left), and the proposed approximation with the recognition model in the form of a bi-RNN (right). Black arrows show conditional dependencies of the model, red arrows show the data-flow in the recognition.}}
\label{fig:gm}
\end{figure}

Our inference scheme uses variational Bayes \citep[see e.g.,][]{beal2003variational, blei2017variational}. We first define the form of the approximation to the posterior, $q(\cdot)$. Then we derive the evidence lower bound (ELBO) with respect to which the posterior approximation is optimised in order to minimise the Kullback-Leibler divergence between the approximate and true posterior. We detail how the ELBO is estimated in a stochastic fashion and optimized using gradient-based methods, and describe how the form of the approximate posterior is given by a recurrent neural network. The graphical models of the GPSSM and our proposed approximation are shown in Figure \ref{fig:gm}.

\subsection{Posterior approximation}
Following the work by \cite{frigola2014variational}, we adopt a variational
approximation to the posterior, assuming factorisation between the
latent functions $f(\cdot)$ and the state trajectories $\vec
X$. However, unlike Frigola et al.'s work, we do not run particle MCMC
to approximate the state trajectories, but instead assume that the posterior
over states is given by a Markov-structured Gaussian distribution
parameterised by a recognition model (see section~\ref{sec:recognition
  model}). In concordance with \citet{frigola2014variational}, we
adopt a sparse variational framework to approximate the GP.  The
sparse approximation allows us to deal with both (a) the unobserved
nature of the GP inputs and (b) any potential computational scaling
issues with the GP by controlling the number of inducing points in the
approximation.

The variational approximation to the GP posterior is formed as follows:
Let $\vec Z = [\vec z_1,\dotsc,\vec z_M]$ be some points in the same domain as $\tilde{\vec x}$. 
For each Gaussian process $f_d(\cdot)$, we define the inducing variables $\vec u_d = [f_d(\vec z_m)]_{m=1}^M$, 
so that the density of $\vec u_d$ under the GP prior is $\mathcal N(\vec \eta_d, \vec K_{zz})$, with 
$\vec \eta_d = [\eta_d(\vec z_m)]_{m=1}^M$. We make a mean-field variational approximation to
the posterior for $\vec U$, taking the form 
$q(\vec U) = \prod_{d=1}^D \mathcal N(\vec u_d\given \vec \mu_d, \vec\Sigma_d)$. 
The variational posterior of the {\em rest}  of the points on the GP is 
assumed to be given by the same conditional distribution as the prior:
\begin{equation}
	f_d(\cdot) \given \vec u_d \sim \mathcal {GP}\big(\eta_d(\cdot) + k(\cdot, \vec Z)\vec K_{zz}^{-1}(\vec u_d - \vec \eta_d), \quad k(\cdot, \cdot) - k(\cdot, \vec Z)\vec K_{zz}^{-1} k(\vec Z, \cdot)\big)\,.
\end{equation}
Integrating this expression with respect to the prior distribution
$p(\vec u_d) = \mathcal N(\vec \eta_d, \,\vec K_{zz})$ gives the GP prior
in eq.~\eqref{eq:prior}.  Integrating with respect to the variational
distribution $q(\vec U)$ gives our approximation to the posterior
process
$f_d(\cdot) \sim \mathcal {GP}\big(\mu_d(\cdot), v_d(\cdot,
\cdot)\big)$, with
\begin{align}
	\label{eq:q_F}
	\mu_d(\cdot) &= \eta_d(\cdot) + k(\cdot, \vec Z)\vec K_{zz}^{-1}(\vec \mu_d - \vec \eta_d),\\ 
		v_d(\cdot, \cdot) &= k(\cdot, \cdot) - k(\cdot, \vec Z)\vec
        K_{zz}^{-1}[\vec K_{zz} - \vec \Sigma_d]\vec K_{zz}^{-1} k(\vec
        Z, \cdot)\,.
	\label{eq:q_F_v}
\end{align}

The approximation to the posterior of the state trajectory is assumed to have a Gauss-Markov structure:
\begin{equation}
	q(\vec x_0) = \mathcal N\big(\vec x_0 \given \vec m_0, \vec L_0 \vec L_0^\top\big),\quad
	q(\vec x_t\given \vec x_{t-1}) = \mathcal N\big(\vec x_t \given \vec A_t \vec x_{t-1}, \vec L_t \vec L_t^\top\big)\,. 
	\label{eq:q_X}
\end{equation}
This distribution is specified through a single mean vector $\vec
m_0$, a series of square matrices $\vec A_t$, and a series of
lower-triangular matrices $\vec L_t$. It serves as a locally linear approximation
to an overall non-linear posterior over the states. 
This is a good approximation provided that the $\Delta t$ between the transitions
is sufficiently small.

With the approximating distributions for the variational posterior
defined in eq.~\eqref{eq:q_F}--\eqref{eq:q_X}, we are ready to derive the evidence lower bound (ELBO) on the model's true likelihood. Following \citep[][eq.~(5.10)]{frigola2015bayesian}, the ELBO is given by
\begin{align}
	\textsc{ELBO} &= \mathbb E _{q(\vec x_0)}[\log p(\vec x_0)] + \textsc H[q(\vec X)] - \textsc{KL}[q(\vec U)\,||\,p(\vec U)]\nonumber\\
	&\quad + \mathbb E_{q(\vec X)} \Big[\sum_{t=1}^T \sum_{d=1}^D-\frac{1}{2\sigma_f^2} v_d(\tilde{\vec x}_{t-1}, \tilde{\vec x}_{t-1}) + \log \mathcal N\big(x_t^{(d)} \given \mu_d(\tilde{\vec x}_{t-1}), \sigma^2_f\big)\Big]\nonumber\\
	&\quad + \mathbb E_{q(\vec X)} \Big[\sum_{t=1}^T \log \mathcal N\big(\vec y_t \given g(\vec x_t), \sigma^2_g\vec I_O \big)\Big]\,,
	\label{eq:ELBO}
\end{align}
where $\textsc{KL}[\cdot||\cdot]$ is the Kullback-Leibler divergence, and $\textsc{H}[\cdot]$ denotes the entropy. Note that with the above formulation we can naturally deal with multiple episodic data since the ELBO can be factorised across independent episodes. We can now learn the GPSSM by optimising the ELBO w.r.t.~the parameters of the model and the variational parameters.  A full derivation is provided in the supplementary material.

The form of the ELBO justifies the Markov-structure that we have assumed for the variational distribution $q(\vec X)$: we see that the latent states only interact over pairwise time steps $\vec x_t$ and $\vec x_{t-1}$; adding further structure to $q(\vec X)$ is unnecessary.

\subsection{Efficient computation of the ELBO}
\label{sec:reparam}
To compute the ELBO in eq.~\eqref{eq:ELBO}, we need to compute expectations w.r.t. 
$q(\vec X)$. \citet{frigola2014variational} showed that for the RBF kernel 
the relevant expectations can be computed in closed form in a similar way to \citet{titsias2010}.
To allow for general kernels we propose to use the reparameterisation 
trick~\citep{kingma2013auto,rezende2014stochastic} instead: by sampling a single trajectory 
from $q(\vec X)$  and evaluating the integrands in eq.~\eqref{eq:ELBO}, 
we obtain an unbiased estimate of the ELBO. To draw a sample from the 
Gauss-Markov structure in eq.~\eqref{eq:q_X}, we first sample 
$\vec \epsilon_t \sim \mathcal N(\vec 0, \vec I),\, t=0,\ldots, T$, 
and then apply recursively the affine transformation
\begin{equation}
	\vec x_0 = \vec m_0 + \vec L_0 \vec \epsilon_0,\quad
	\vec x_t = \vec A_t \vec x_{t-1} + \vec L_t \vec \epsilon_t\,.
\end{equation}
This simple estimator of the ELBO can then be used in optimisation using 
stochastic gradient methods; we used the Adam optimizer~\citep{kingma2014adam}.
It may seem initially counter-intuitive to use a stochastic estimate
of the ELBO where one is available in closed form, 
but this approach offers two distinct advantages.
First, computation is dramatically reduced: our scheme requires $\mathcal O(TD)$ 
storage in order to evaluate the integrand in eq.~\eqref{eq:ELBO} 
at a single sample from $q(\vec X)$. A scheme that computes the integral in closed 
form requires $\mathcal O(TM^2)$ (where M is the number of inducing variables in the sparse GP)
storage for the sufficient statistics of the kernel evaluations. 
The second advantage is that we are no longer restricted to the RBF kernel, 
but can use any valid kernel for inference and learning in GPSSMs. 
The reparameterisation trick also allows us to perform batched updates of the model parameters,
amounting to doubly stochastic variational inference \citep{Titsias2014}, 
which we experimentally found to improve run-time and sample-efficiency.

Some of the elements of the ELBO in eq.~\eqref{eq:ELBO} are still available
in closed-form.  To reduce the variance of the estimate of the ELBO we exploit 
this where possible: the entropy of the Gauss-Markov structure is 
$\textsc{H}[q(\vec X)] = -\frac{TD}{2}\log (2 \pi e) - \sum_{t=0}^T\log(\det(\vec L_t))$; 
the expected likelihood (last term in eq.~\eqref{eq:ELBO}) can be computed 
easily given the marginals of $q(\vec X)$, which are given by
\begin{align}
	q(\vec x_t) = \mathcal N(\vec m_t, \vec \Sigma_t),\quad
	\vec m_t = \vec A_t \vec m_{t-1},\quad \vec \Sigma_t = \vec A_t \vec \Sigma_{t-1} \vec A_t^\top + \vec L_t\vec L_t^\top\,,
	\label{eq:state-marginals}
\end{align}
and the necessary Kullback-Leibler divergences can be computed analytically:
we use the implementations from GPflow \citep{matthews2017gpflow}.

\subsection{A recurrent recognition model}
\label{sec:recognition model}

The variational distribution of the latent trajectories in eq.~\eqref{eq:q_X} has a large number of parameters ($\vec A_t, \vec L_t$) that grows with the length of the dataset. Further, if we wish to train a model on multiple episodes (independent data sequences sharing the same dynamics), then the number of parameters grows further. To alleviate this, we propose to use a recognition model in the form of a bi-directional recurrent neural network (bi-RNN), which is responsible for recovering the variational parameters $\vec A_t, \vec L_t$.

A bi-RNN is a combination of two independent RNNs operating on opposite directions of the sequence.  Each network is specified by two weight matrices $\vec W$ acting on a hidden state $\vec h$:
\begin{align}
	\vec h_{t}^{(f)} &= \phi(\vec W_h^{(f)} \vec h_{t-1}^{(f)} + \vec W_{\tilde{y}}^{(f)} \tilde{\vec y}_{t} + \vec b_h^{(f)} )\,, \quad \textrm{forward passing}\\
	\vec h_{t}^{(b)} &= \phi(\vec W_h^{(b)} \vec h_{t+1}^{(b)} + \vec W_{\tilde{y}}^{(b)} \tilde{\vec y}_t + \vec b_h^{(b)} )\,, \quad \textrm{backward passing}
	\label{eq:rnn}
\end{align}
where $\tilde{\vec y}_t = [\vec y_t, \vec a_t]$ denotes the concatenation of the observed data and control actions and the superscripts denote the direction (forward/backward) of the RNN. The activation function $\phi$ (we use the $\tanh$ function), acts on each element of its argument separately. In our experiments we found that using gated recurrent units \citep{Cho14} improved performance of our model. We now make the parameters of the Gauss-Markov structure dependent on the sequences $\vec h^{(f)}, \vec h^{(b)}$, so that
\begin{align}
	\vec A_t = \textrm{reshape}(\vec W_{\!\!A} [\vec h_t^{(f)}; \vec h_t^{(b)}] + \vec b_A),\quad \vec L_t = \textrm{reshape}(\vec W_{\!\!L} [\vec h_t^{(f)}; \vec h_t^{(b)}]  + \vec b_L) \,.
	\label{eq:recognition}
\end{align}

The parameters of the Gauss-Markov structure $q(\vec X)$ are now almost completely encapsulated in the recurrent recognition model as $\vec W_h^{(f,b)}, \vec W_{\tilde{y}}^{(f,b)}, \vec W_{\!\!A}, \vec W_{\!\!L}, \vec b_h^{(f,b)}, \vec b_A, \vec b_L$. We only need to infer the parameters of the initial state, $\vec m_0, \vec L_0$ for each episode; this is where we utilise the functionality of the bi-RNN structure. Instead of directly learning the initial state $q(\vec{x}_0)$, we can now obtain it indirectly via the output state of the backward RNN. Another nice property of the proposed recognition model is that now $q(\vec X)$ is recognised from both future and past observations, since the proposed bi-RNN recognition model can be regarded as a forward and backward sequential smoother of our variational posterior. Finally, it is worth noting the interplay between the variational distribution $q(\vec X)$ and the recognition model.
Recall that the variational distribution is a Bayesian linear approximation to the non-linear posterior
and is fully defined by the time varying parameters, $\vec A_t, \vec L_t$; the recognition model has the
role to recover these parameters via the non-linear and time invariant RNN.




\definecolor{color0}{rgb}{0.12156862745098,0.466666666666667,0.705882352941177}
\definecolor{color1}{rgb}{1,0.498039215686275,0.0549019607843137}
\section{Experiments}
We benchmark the proposed GPSSM approach on data from
one illustrative example and three challenging non-linear data sets of simulated and real data. Our aim is to
demonstrate that we can: (i) benefit from the use of non-smooth
    kernels with our approximate inference and accurately model
    non-smooth transition functions;
(ii) successfully learn
    non-linear dynamical systems even from noisy and partially observed inputs;
    (iii) sample plausible future
    trajectories from the system
    even when trained with either a small number of episodes or long time sequences.
    
\subsection{Non-linear system identification}
We first apply our approach to a synthetic dataset generated broadly according to \citep{frigola2014variational}. 
The data is created using a non-linear, non-smooth transition function with additive state and observation noise 
according to: $p(x_{t + 1} | x_t) = \mathcal{N}(f(x_t), \sigma_f^2)$, and $p(y_t | x_t) = \mathcal{N}(x_t, \sigma_g^2)$, where
\begin{equation}
\label{eq:one_d_transition}
f(x_t) = x_{t} + 1, \quad \textrm{if } x_t < 4, \qquad 13 - 2 x_t, \quad \textrm{otherwise}\,.
\end{equation}
In our experiments, we set the system and measurement noise variances to $\sigma_f^2 = 0.01$
and $\sigma_g^2 = 0.1$, respectively, and generate 200 episodes of
length  $10$ that were used as the observed data for training the GPSSM. We used $20$ inducing points 
(initialised uniformly across the range of the input data) for approximating the GP and 20 hidden units
for the recurrent recognition model. We evaluate the following kernels:
RBF, additive composition of the RBF (initial $\ell=10$) and Matern
($\nu = \frac{1}{2}$, initial $\ell=0.1$),
$0$-order arc-cosine~\citep{Cho09}, and the MGP
kernel~\citep{Calandra2016} (depth 5, hidden dimensions $[3, 2, 3, 2,
3]$, $\tanh$ activation, Matern ($\nu = \frac{1}{2}$) compound
kernel).

\begin{figure}
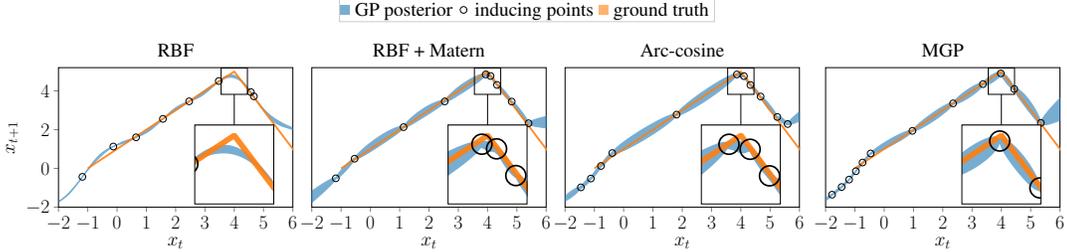

\centering
\setlength{\tabcolsep}{.01pt}
\begin{tabular}{lCLl}
\scalebox{.42}{\input{\figpath/one_d/RBF_0.1_0.01_transition_model.tex}} &
\scalebox{.42}{\input{\figpath/one_d/RBFMatern_0.1_0.01_transition_model.tex}} &
\scalebox{.42}{\input{\figpath/one_d/ArcCosine0_0.1_0.01_transition_model.tex}} &
\scalebox{.42}{\input{\figpath/one_d/MGP_0.1_0.01_transition_model.tex}}
\end{tabular}
\caption{{\small The learnt state transition function with different kernels. The true function is given by eq. \eqref{eq:one_d_transition}.}}
\label{fig:one_d}
\end{figure}

The learnt GP state transition functions are shown in
Figure~\ref{fig:one_d}. With the non-smooth kernels
we are able to learn accurate transitions and model the instantaneous
dynamical change, as opposed to the smooth transition learnt with the
RBF. Note that all non-smooth kernels place inducing points directly
on the peak (at $x_t=4$) to model the kink, whereas the RBF kernel
explains this behaviour as a longer-scale wiggliness of the posterior
process. When using a kernel without
the RBF component the GP posterior quickly reverts to the mean
function ($\eta(x) = x$) as we move away from the data: the short
length-scales that enable them to model the instantaneous change
prevent them from extrapolating downwards in the transition
function. The composition of the RBF and Matern kernel benefits from long and short length scales and can better extrapolate. The posteriors can be viewed across a longer range of the function space in the supplementary material.

\setlength\figureheight{6cm}
\setlength\figurewidth{9cm}

\newcolumntype{C}{@{\extracolsep{-5.6cm}}l@{\extracolsep{-1pt}}}%
\newcolumntype{L}{@{\extracolsep{-.1cm}}l@{\extracolsep{-.1pt}}}%

\subsection{Modelling cart-pole dynamics}\label{sec:cart_pole}
We demonstrate the efficacy of the proposed GPSSM
on learning the non-linear dynamics of the cart-pole system
from~\citep{deisenroth2011pilco}. The system is composed of a cart
running on a track, with a freely swinging pendulum attached to
it. The state of the system consists of the cart's position and
velocity, and the pendulum's angle and angular velocity, while a
horizontal force (action) $a \in [-10, 10] N$ can be applied to the
cart.  We used the PILCO algorithm
from~\citep{deisenroth2011pilco} to learn a feedback controller that
swings the pendulum and balances it in the inverted position in
the middle of the track.  We collected trajectory data from 16 trials during
learning; each trajectory/episode  was $\unit[4]{s}$ (40 time steps)
long. 

When training the GPSSM for the cart-pole system we used data up to the first 15 episodes. We used $100$ inducing points to approximate the GP function with a Matern $\nu=\frac{1}{2}$ and $50$ hidden units for the recurrent recognition model. The learning rate for the Adam optimiser was set to $10^{-3}$.
 We qualitatively assess the performance of our model by feeding the
control sequence of the last episode to the GPSSM in order to generate future responses.

\begin{figure}
\centering
\scalebox{.52}{\input{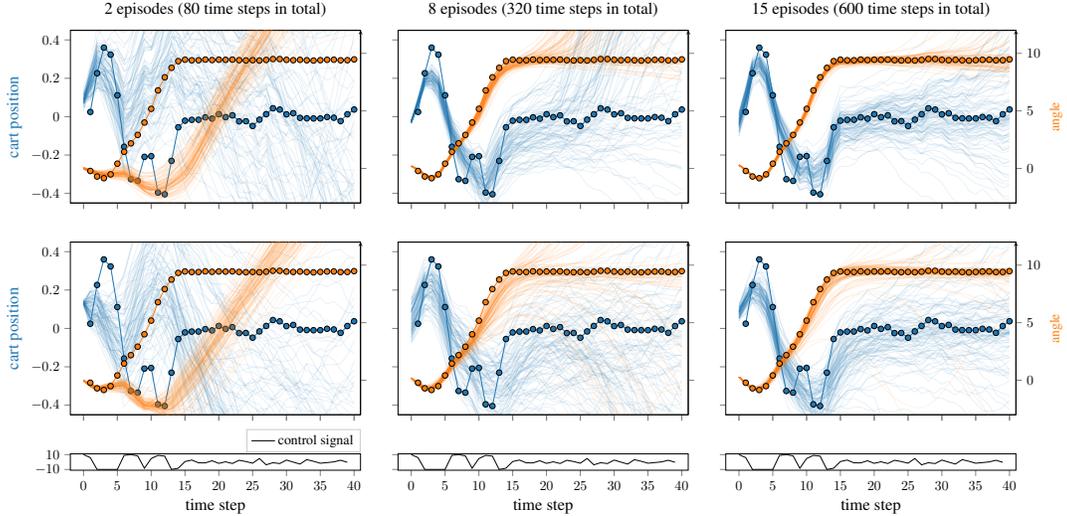}}
\caption{{\small Predicting the \textcolor{color0}{cart's position} and \textcolor{color1}{pendulum's angle} behaviour from the cart-pole dataset by applying the control signal of the testing episode to sampled future trajectories from the proposed GPSSM. Learning of the dynamics is demonstrated with \emph{observed} (upper row) and \emph{hidden} (lower row) velocities and with increasing number of training episodes. Ground truth is denoted with the marked lines.}}
\label{fig:samples_cart_pole}
\end{figure}

In Figure~\ref{fig:samples_cart_pole}, we demonstrate the ability of
the proposed GPSSM to learn the underlying dynamics of the system from
a different number of episodes with
fully and partially observed data. In the top row, the GPSSM
observes the full 4D state, while in the bottom row, we train the GPSSM with only the cart's position
and the pendulum's angle observed (i.e., the true state is not fully observed since the velocities are hidden).
In both cases, sampling long-term trajectories based on only 2
episodes for training does not result in plausible future
trajectories. However, we could  model part of the dynamics after
training with only 8 episodes (320 time steps interaction with the
system), while training with 15 episodes (600 time steps in total)
allowed the GPSSM to produce trajectories similar to the
ground truth. It is worth emphasising the fact that the GPSSM could recover the unobserved velocities in the latent states, which
resulted in smooth transitions of the cart and swinging of the
pendulum. However, it seems that the recovered cart's velocity is overestimated. This is evidenced by the increased variance
in the prediction of the cart's position around $0$ (the centre of the track).
Detailed fittings for each episode and learnt latent states with observed and hidden velocities are provided in the supplementary material.

\begin{table}[!h]
	\centering
	\begin{varwidth}{.5\linewidth}
		\small
		\caption{{\small Average Euclidean distance between the true and the predicted trajectories, measured at the pendulum's tip. The error is in pendulum's length units.}}
		\label{tab:rmse}
		\begin{tabular}{cccc}
			\toprule
			& \textrm{2 episodes} & \textrm{8 episodes} & \textrm{15 episodes}\\
			\midrule
			\textrm{Kalman} & 1.65 & 1.52 & 1.48\\
			\textrm{ARGP} & {\bf1.22} & 1.03 & 0.80\\
			\textrm{GPSSM} & 1.21 & {\bf 0.67} & {\bf 0.59}\\
			\bottomrule
		\end{tabular}
	\end{varwidth}%
	\begin{minipage}{.5\linewidth}
		\centering
		\scalebox{.52}{\input{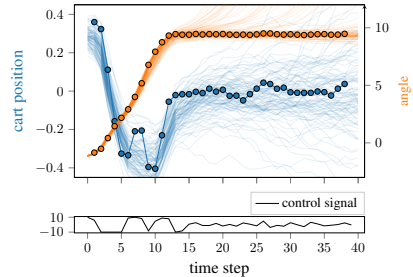}}
		\captionof{figure}{{\small Predictions with lagged actions.}}
		\label{fig:lagged_actions}
	\end{minipage}
\end{table}

In Table~\ref{tab:rmse}, we provide the average Euclidean distance between the predicted and the true trajectories measured at the pendulum's tip, with fully observed states.
We compare to two baselines:
(i) the auto-regressive GP (ARGP) that maps the tuple $[\vec y_{t-1}, a_{t-1}]$ to the next observation $\vec y_t$ (as in PILCO \citep{Deisenroth2015}), 
and (ii) a linear system for identification that uses the Kalman filtering technique~\citep{Kalman1960}.
We see that the GPSSM significantly outperforms the baselines on this highly non-linear benchmark.
The linear system cannot learn the dynamics at all, while the ARGP only manages to produce sensible error
(less than a pendulum's length) after seeing 15 episodes. Note that the GPSSM trained on 8 episodes
produces trajectories with less error than the ARGP trained on 15 episodes.

We also ran experiments using lagged actions where the partially observed state at time $t$ is affected by the action at $t-2$.
Figure~\ref{fig:lagged_actions} shows that we are able to sample future trajectories with an accuracy similar to time-aligned actions.
This indicates that our model is able to learn a compressed representation of the full state and previous inputs, essentially `remembering' the lagged actions.

\subsection{Modelling double pendulum dynamics}\label{sec:double-pendulum}

We demonstrate the learning and
modelling of the dynamics of the double pendulum system
from~\citep{Deisenroth2015}. The double pendulum is a two-link
robot arm with two actuators. The state of the system consists of the angles and the corresponding angular 
velocities of the inner and outer link, respectively, while different torques $a_1, a_2 \in [-2, 2]\,\text{Nm}$ 
can be applied to the two actuators. The task of swinging the double pendulum and balancing it in the
upwards position is extremely challenging. First, it requires the interplay of two correlated control
signals (i.e., the torques). Second, the behaviour of the system, when operating at free will, is chaotic.

\begin{figure}[thb]
\centering
\scalebox{.52}{
\input{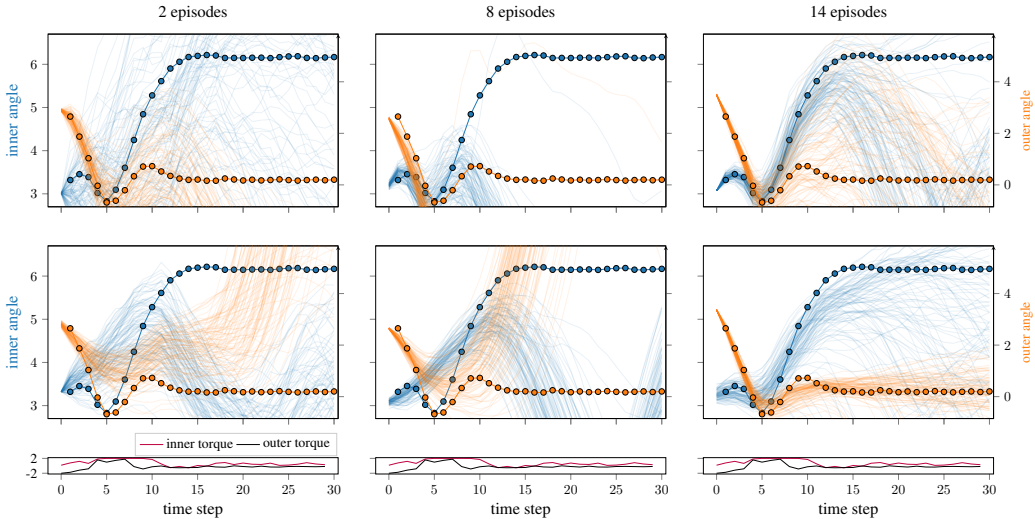}
}
\caption{Predicting the \textcolor{color0}{inner} and \textcolor{color1}{outer} pendulum's angle from the double pendulum dataset by applying the control signals of the testing episode to sampled future trajectories from the proposed GPSSM. Learning of the dynamics is demonstrated with \emph{observed} (upper row) and \emph{hidden} (lower row) angular velocities and with increasing number of training episodes. Ground truth is denoted with the marked lines.}
	\label{fig:samples_double_pendulum}
\end{figure}

We learn the underlying dynamics from episodic data (15 episodes, 30 time
steps long each). Training of the GPSSM was performed with data up to
14 episodes, while always demonstrating the learnt underlying dynamics
on the last episode, which serves as the test set. We used $200$ inducing points to approximate 
the GP function with a Matern $\nu=\frac{1}{2}$ and $80$ hidden units for the recurrent recognition 
model. The learning rate for the Adam optimiser was set to $10^{-3}$.
The difficulty of the task is evident in Figure~\ref{fig:samples_double_pendulum}, where we 
can see that even after observing 14 episodes we cannot accurately predict the system's future 
behaviour for more than 15 time steps (i.e., $\unit[1.5]{s}$). 
It is worth noting that we can generate reliable simulation even though we observe only the pendulums' angles.


\subsection{Modelling actuator dynamics}
\begin{figure}
\centering
	\scalebox{.52}{\input{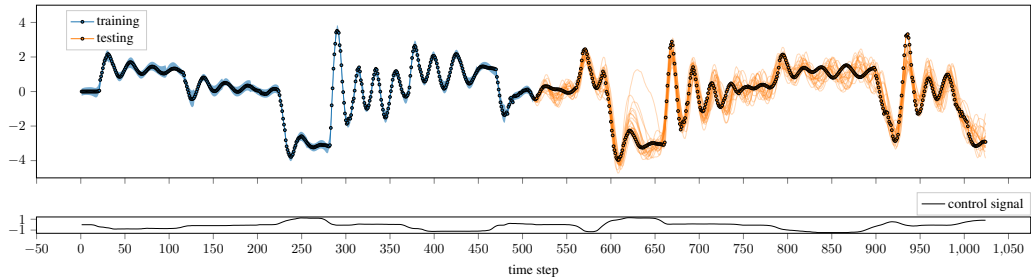}}
	\caption{{\small Demonstration of the identified model that controls the non-linear dynamics of the actuator dataset.
	The model's fitting on the \textcolor{color0}{train data} and sampled \textcolor{color1}{future predictions}, after applying the control signal to the system.  Ground truth is denoted with the marked lines.}} 
	\label{fig:actuator}
\end{figure}

Here we evaluate the proposed GPSSM on real data from a hydraulic
actuator that controls a robot arm \citep{sjoberg1995nonlinear}. The input is the size of the actuator’s valve opening and the
output is its oil pressure. We train the GPSSM on half the sequence (512 steps) and evaluate the model on the remaining half.
We use 15 inducing points to approximate the GP function with a 
combination of an RBF and a Matern $\nu=\frac{1}{2}$ 
and 15 hidden units for the recurrent recognition model.
Figure~\ref{fig:actuator} shows the fitting on the train data 
along with sampled future predictions from the learnt system when
operating on a free simulation mode.
It is worth noting the correct capturing of the uncertainty from the model at the points
where the predictions are not accurate.

\section{Discussion and conclusion}

We have proposed a novel inference mechanism for the GPSSM, in order to address the challenging
task of non-linear system identification. Since our inference is based on 
the variational framework, successful learning of the model 
relies on defining good approximations to 
the posterior of the latent functions and states.
Approximating the posterior over the dynamics with a sparse GP seems to be a reasonable
choice given our assumptions over the transition function. However, the difficulty remains in
the selection of the approximate posterior of the latent states. This is the key component
that enables successful learning of the GPSSM.
 
In this work, we construct the variational posterior so that it follows the same
Markov properties as the true states. Furthermore, it is enforced to have 
a simple-to-learn, linear, time-varying structure.
To assure, though, that this approximation has rich representational
capacity we proposed to recover the variational parameters of the posterior via a non-linear
recurrent recognition model. Consequently, the joint approximate posterior resembles the
behaviour of the true system, which facilitates the effective learning of the GPSSM.

In the experimental section we have provided evidence that the proposed approach is able to 
identify latent dynamics in true and simulated data, even from partial and lagged observations,
 while requiring only small data sets for this challenging task.


\subsection*{Acknowledgement}
Marc P.~Deisenroth has been supported by a Google faculty research award.
\bibliographystyle{plainnat}
\bibliography{references}

\begin{thebibliography}{40}
\providecommand{\natexlab}[1]{#1}
\providecommand{\url}[1]{\texttt{#1}}
\expandafter\ifx\csname urlstyle\endcsname\relax
  \providecommand{\doi}[1]{doi: #1}\else
  \providecommand{\doi}{doi: \begingroup \urlstyle{rm}\Url}\fi

\bibitem[Al-Shedivat et~al.(2016)Al-Shedivat, Wilson, Saatchi, Hu, and
  Xing]{al2016learning}
Maruan Al-Shedivat, Andrew~G. Wilson, Yunus Saatchi, Zhiting Hu, and Eric~P.
  Xing.
\newblock Learning scalable deep kernels with recurrent structure.
\newblock \emph{arXiv preprint arXiv:1610.08936}, 2016.

\bibitem[Beal(2003)]{beal2003variational}
Matthew~J. Beal.
\newblock \emph{Variational algorithms for approximate Bayesian inference}.
\newblock PhD thesis, University of London, London, UK, 2003.

\bibitem[Blei et~al.(2017)Blei, Kucukelbir, and McAuliffe]{blei2017variational}
David~M. Blei, Alp Kucukelbir, and Jon~D. McAuliffe.
\newblock Variational inference: A review for statisticians.
\newblock \emph{Journal of the American Statistical Association}, 112\penalty0
  (518):\penalty0 859--877, 2017.

\bibitem[Brown et~al.(1998)Brown, Frank, Tang, Quirk, and
  Wilson]{brown1998statistical}
Emery~N. Brown, Loren~M. Frank, Dengda Tang, Michael~C. Quirk, and Matthew~A.
  Wilson.
\newblock A statistical paradigm for neural spike train decoding applied to
  position prediction from ensemble firing patterns of rat hippocampal place
  cells.
\newblock \emph{Journal of Neuroscience}, 18\penalty0 (18):\penalty0
  7411--7425, 1998.

\bibitem[Calandra et~al.(2016)Calandra, Peters, Rasmussen, and
  Deisenroth]{Calandra2016}
Roberto Calandra, Jan Peters, Carl~E. Rasmussen, and Marc~P. Deisenroth.
\newblock {Manifold Gaussian processes for regression}.
\newblock In \emph{IEEE International Joint Conference on Neural Networks},
  2016.

\bibitem[Cho et~al.(2014)Cho, van Merrienboer, Bahdanau, and Bengio]{Cho14}
KyungHyun Cho, Bart van Merrienboer, Dzmitry Bahdanau, and Yoshua Bengio.
\newblock On the properties of neural machine translation: Encoder-decoder
  approaches.
\newblock \emph{arXiv preprint arXiv:1409.1259}, 2014.

\bibitem[Cho and Saul(2009)]{Cho09}
Youngmin Cho and Lawrence~K. Saul.
\newblock Kernel methods for deep learning.
\newblock In \emph{Advances in Neural Information Processing Systems}, pages
  342--350. 2009.

\bibitem[Dai et~al.(2015)Dai, Damianou, Gonz{\'a}lez, and
  Lawrence]{dai2015variational}
Zhenwen Dai, Andreas Damianou, Javier Gonz{\'a}lez, and Neil Lawrence.
\newblock Variational auto-encoded deep {G}aussian processes.
\newblock In \emph{International Conference on Learning Representations}, 2015.

\bibitem[Deisenroth and Mohamed(2012)]{Deisenroth2012d}
Marc~P. Deisenroth and Shakir Mohamed.
\newblock {Expectation {propagation in Gaussian process dynamical systems}}.
\newblock In \emph{{Advances in Neural Information Processing Systems}}, pages
  2618--2626, 2012.

\bibitem[Deisenroth and Rasmussen(2011)]{deisenroth2011pilco}
Marc~P. Deisenroth and Carl~E. Rasmussen.
\newblock {PILCO}: A model-based and data-efficient approach to policy search.
\newblock In \emph{International Conference on Machine Learning}, pages
  465--472, 2011.

\bibitem[Deisenroth et~al.(2009)Deisenroth, Huber, and
  Hanebeck]{Deisenroth2009a}
Marc~P. Deisenroth, Marco~F. Huber, and Uwe~D. Hanebeck.
\newblock {Analytic {moment-based Gaussian process filtering}}.
\newblock In \emph{{International Conference on Machine Learning}}, pages
  225--232, 2009.

\bibitem[Deisenroth et~al.(2012)Deisenroth, Turner, Huber, Hanebeck, and
  Rasmussen]{Deisenroth2012}
Marc~P. Deisenroth, Ryan~D. Turner, Marco Huber, Uwe~D. Hanebeck, and Carl~E.
  Rasmussen.
\newblock {Robust {filtering and smoothing with Gaussian processes}}.
\newblock \emph{IEEE Transactions on Automatic Control}, 57\penalty0
  (7):\penalty0 1865--1871, 2012.

\bibitem[Deisenroth et~al.(2015)Deisenroth, Fox, and Rasmussen]{Deisenroth2015}
Marc~P. Deisenroth, Dieter Fox, and Carl~E. Rasmussen.
\newblock {Gaussian {processes for data-efficient learning in robotics and
  control}}.
\newblock \emph{IEEE Transactions on Pattern Analysis and Machine
  Intelligence}, 37\penalty0 (2):\penalty0 408--423, 2015.

\bibitem[Frigola(2015)]{frigola2015bayesian}
Roger Frigola.
\newblock \emph{Bayesian time series learning with {G}aussian processes}.
\newblock PhD thesis, University of Cambridge, Cambridge, UK, 2015.

\bibitem[Frigola et~al.(2013)Frigola, Lindsten, Sch{\"o}n, and
  Rasmussen]{frigola2013bayesian}
Roger Frigola, Fredrik Lindsten, Thomas~B. Sch{\"o}n, and Carl~E. Rasmussen.
\newblock Bayesian inference and learning in {G}aussian process state-space
  models with particle {MCMC}.
\newblock In \emph{Advances in Neural Information Processing Systems}, pages
  3156--3164, 2013.

\bibitem[Frigola et~al.(2014)Frigola, Chen, and
  Rasmussen]{frigola2014variational}
Roger Frigola, Yutian Chen, and Carl~E. Rasmussen.
\newblock Variational {G}aussian process state-space models.
\newblock In \emph{Advances in Neural Information Processing Systems}, pages
  3680--3688, 2014.

\bibitem[Kalman(1960)]{Kalman1960}
Rudolf~E. Kalman.
\newblock {A new approach to linear filtering and prediction problems}.
\newblock \emph{Transactions of the American Society of Mathematical
  Engineering, Journal of Basic Engineering}, 82\penalty0 (D):\penalty0 35--45,
  1960.

\bibitem[Kingma and Ba(2015)]{kingma2014adam}
Diederik~P. Kingma and Jimmy Ba.
\newblock Adam: A method for stochastic optimization.
\newblock In \emph{International Conference on Learning Representations}, 2015.

\bibitem[Kingma and Welling(2014)]{kingma2013auto}
Diederik~P. Kingma and Max Welling.
\newblock Auto-encoding variational {B}ayes.
\newblock In \emph{International Conference on Learning Representations}, 2014.

\bibitem[Ko and Fox(2009{\natexlab{a}})]{Ko2009}
Jonathan Ko and Dieter Fox.
\newblock {G{P-BayesFilters: Bayesian filtering using Gaussian process
  prediction and observation models}}.
\newblock \emph{Autonomous Robots}, 27\penalty0 (1):\penalty0 75--90,
  2009{\natexlab{a}}.

\bibitem[Ko and Fox(2009{\natexlab{b}})]{Ko2009a}
Jonathan Ko and Dieter Fox.
\newblock {Learning {GP}-{B}ayes{F}ilters via {G}aussian {process latent
  variable models}}.
\newblock In \emph{{Robotics: Science and Systems}}, 2009{\natexlab{b}}.

\bibitem[Kocijan(2016)]{kocijan2016modelling}
Ju{\v{s}} Kocijan.
\newblock \emph{Modelling and control of dynamic systems using {G}aussian
  process models}.
\newblock Springer, 2016.

\bibitem[Likar and Kocijan(2007)]{likar2007predictive}
Bojan Likar and Ju{\v{s}} Kocijan.
\newblock Predictive control of a gas-liquid separation plant based on a
  {G}aussian process model.
\newblock \emph{Computers \& Chemical Engineering}, 31\penalty0 (3):\penalty0
  142--152, 2007.

\bibitem[Ljung(1999)]{Ljung1999}
Lennart Ljung.
\newblock \emph{{System {identification: Theory for the user}}}.
\newblock Prentice Hall, 1999.

\bibitem[Matthews(2017)]{matthews2017thesis}
Alexander G. de~G. Matthews.
\newblock \emph{Scalable Gaussian process inference using variational methods}.
\newblock PhD thesis, University of Cambridge, Cambridge, UK, 2017.

\bibitem[Matthews et~al.(2016)Matthews, Hensman, Turner, and
  Ghahramani]{matthews2016sparse}
Alexander G. de~G. Matthews, James Hensman, Richard~E. Turner, and Zoubin
  Ghahramani.
\newblock On sparse variational methods and the {K}ullback-{L}eibler divergence
  between stochastic processes.
\newblock In \emph{International Conference on Artificial Intelligence and
  Statistics}, volume~51 of \emph{{JMLR W\&CP}}, pages 231--239, 2016.

\bibitem[Matthews et~al.(2017)Matthews, van~der Wilk, Nickson, Fujii,
  Boukouvalas, Le{\'o}n-Villagr{\'a}, Ghahramani, and
  Hensman]{matthews2017gpflow}
Alexander G. de~G. Matthews, Mark van~der Wilk, Tom Nickson, Keisuke Fujii,
  Alexis Boukouvalas, Pablo Le{\'o}n-Villagr{\'a}, Zoubin Ghahramani, and James
  Hensman.
\newblock G{P}flow: A {G}aussian process library using {T}ensor{F}low.
\newblock \emph{Journal of Machine Learning Research}, 18\penalty0
  (40):\penalty0 1--6, 2017.

\bibitem[Mattos et~al.(2015)Mattos, Dai, Damianou, Forth, Barreto, and
  Lawrence]{mattos2015recurrent}
C{\'e}sar Lincoln~C. Mattos, Zhenwen Dai, Andreas Damianou, Jeremy Forth,
  Guilherme~A. Barreto, and Neil~D. Lawrence.
\newblock Recurrent {G}aussian processes.
\newblock In \emph{International Conference on Learning Representations}, 2015.

\bibitem[Murray-Smith and Girard(2001)]{murray2001gaussian}
Roderick Murray-Smith and Agathe Girard.
\newblock Gaussian process priors with {ARMA} noise models.
\newblock In \emph{Irish Signals and Systems Conference}, pages 147--152, 2001.

\bibitem[Rasmussen and Williams(2006)]{Rasmussen2006}
Carl~E. Rasmussen and Christopher K.~I. Williams.
\newblock \emph{{Gaussian processes for machine learning}}.
\newblock The MIT Press, Cambridge, MA, USA, 2006.

\bibitem[Rezende et~al.(2014)Rezende, Mohamed, and
  Wierstra]{rezende2014stochastic}
Danilo~J. Rezende, Shakir Mohamed, and Daan Wierstra.
\newblock Stochastic backpropagation and approximate inference in deep
  generative models.
\newblock In \emph{International Conference on Machine Learning}, pages
  1278--1286, 2014.

\bibitem[Roberts et~al.(2013)Roberts, Osborne, Ebden, Reece, Gibson, and
  Aigrain]{roberts2013gaussian}
Stephen Roberts, Michael Osborne, Mark Ebden, Steven Reece, Neale Gibson, and
  Suzanne Aigrain.
\newblock Gaussian processes for time-series modelling.
\newblock \emph{Philosophical Transactions of the Royal Society A},
  371\penalty0 (1984):\penalty0 20110550, 2013.

\bibitem[Schneider(1997)]{Schneider1997}
Jeff~G. Schneider.
\newblock {Exploiting {model uncertainty estimates for safe dynamic control
  learning}}.
\newblock In \emph{{Advances in Neural Information Processing Systems}}. 1997.

\bibitem[Sj{\"o}berg et~al.(1995)Sj{\"o}berg, Zhang, Ljung, Benveniste, Delyon,
  Glorennec, Hjalmarsson, and Juditsky]{sjoberg1995nonlinear}
Jonas Sj{\"o}berg, Qinghua Zhang, Lennart Ljung, Albert Benveniste, Bernard
  Delyon, Pierre-Yves Glorennec, H{\aa}kan Hjalmarsson, and Anatoli Juditsky.
\newblock Nonlinear black-box modeling in system identification: A unified
  overview.
\newblock \emph{Automatica}, 31\penalty0 (12):\penalty0 1691--1724, 1995.

\bibitem[Titsias(2009)]{titsias2009variational}
Michalis~K. Titsias.
\newblock Variational learning of inducing variables in sparse {G}aussian
  processes.
\newblock In \emph{{International Conference on Artificial Intelligence and
  Statistics}}, volume~5 of \emph{{JMLR W\&CP}}, pages 567--574, 2009.

\bibitem[Titsias and Lawrence(2010)]{titsias2010}
Michalis~K. Titsias and Neil~D. Lawrence.
\newblock {Bayesian {Gaussian process latent variable model}}.
\newblock In \emph{{International Conference on Artificial Intelligence and
  Statistics}}, volume~9 of \emph{{JMLR W\&CP}}, pages 844--851, 2010.

\bibitem[Titsias and L{\'a}zaro-Gredilla(2014)]{Titsias2014}
Michalis~K. Titsias and Miguel L{\'a}zaro-Gredilla.
\newblock Doubly stochastic variational {B}ayes for non-conjugate inference.
\newblock In \emph{International Conference on Machine Learning}, pages
  1971--1979, 2014.

\bibitem[Turner(2011)]{Turner2011}
Ryan~D. Turner.
\newblock \emph{Gaussian processes for state space models and change point
  detection}.
\newblock PhD thesis, University of Cambridge, Cambridge, UK, 2011.

\bibitem[Turner et~al.(2010)Turner, Deisenroth, and Rasmussen]{Turner2010}
Ryan~D. Turner, Marc~P. Deisenroth, and Carl~E. Rasmussen.
\newblock {State-{space inference and learning with Gaussian processes}}.
\newblock In \emph{{International Conference on Artificial Intelligence and
  Statistics}}, volume~9 of \emph{{JMLR W\&CP}}, pages 868--875, 2010.

\bibitem[Wang et~al.(2008)Wang, Fleet, and Hertzmann]{Wang2008}
Jack~M. Wang, David~J. Fleet, and Aaron Hertzmann.
\newblock {Gaussian {process dynamical models for human motion}}.
\newblock \emph{IEEE Transactions on Pattern Analysis and Machine
  Intelligence}, 30\penalty0 (2):\penalty0 283--298, 2008.

\end{thebibliography}

\clearpage
\begin{appendices}
\section{Derivation of the ELBO}
This appendix contains three parts: we first explicate the joint distribution of the model and data $p(\vec X, \vec f(\cdot), \vec Y)$; then we describe the variational approximation to the model posterior $q(\vec X, \vec f(\cdot))$; then we show how they combine to produce the ELBO. Table \ref{tab:nom} provides some nomenclature.
\begin{table}[h]
\caption{Nomenclature used in this derivation}
\label{tab:nom}
\begin{tabular}{ll}
\toprule
$t\in\{0\ldots T\}$ & time steps  indexed $t$\\
$d\in\{1\ldots D\}$ & dimension of hidden states $\vec x_t$ indexed $d$\\
$O$ & dimension of the observed data\\
$m\in\{1\ldots M\}$ &number of inducing variables indexed $m$\\
\midrule
$\vec x_t$ & hidden state at time $t$, $\vec x_t \in \mathbb R^D$\\
$\vec a_t$ & control input (action) at time $t$, $\vec a_t \in \mathbb R^P$\\
$\tilde{\vec x}_t$ & concatenation of control input and state at $t$\\
$\vec y_t$ & observation at time $t$, $\vec y_t \in \mathbb R^O$\\
$\tilde{\vec y}_t$ & concatenation of control input and observation at $t$\\
\midrule
$\vec X$ & collection of hidden states, $\vec X =[\vec x_t]_{t=0}^T$.\\
$\vec Y$ & collection of observations, $\vec Y =[\vec y_t]_{t=0}^T$.\\
\midrule
$\sigma^2_f$ & variance of state transition noise\\
$\sigma^2_g$ & variance of observation nosie\\
\midrule
$f_d(\cdot)$ & the $d^\textrm{th}$ Gaussian process (GP)\\
$\vec f(\cdot)$ & collection of GPs, $=[f_d(\cdot)]_{d=1}^D$\\
$\eta_d(\cdot)$ & prior mean function of the $d^\textrm{th}$ GP\\
$k_d(\cdot\,, \cdot)$ & prior covariance function of the $d^\textrm{th}$ GP\\
$\mu_d(\cdot)$ & posterior mean function of the $d^\textrm{th}$ GP\\
$v_d(\cdot,\cdot)$ & posterior covariance function of the $d^\textrm{th}$ GP\\
\midrule
$\vec Z$ & Locations of variational pseudo-inputs\\
$\vec u_d$ & evaluations of the $d^\textrm{th}$ GP at the pseudo-inputs: $\vec u_d = [f_d(\vec z_m)]_{m=1}^M$.\\
$\vec U$ & collection: $\vec U = [\vec u_d]_{d=1}^D$\\
$\vec \mu_d$ & variational posterior mean of $\vec u_d$: $\vec \mu_d = [\mu_d(\vec z_m)]_{m=1}^M$\\
$\vec \Sigma_d$ & variational posterior covariance of $\vec u_d$\\
\midrule
$\vec A_t$ & variational transition matrix of $q(\vec x_t\given \vec x_{t-1})$\\
$\vec L_t$ & triangular-square-root of variational covariance of $q(\vec x_t\given \vec x_{t-1})$\\
$\vec m_t$ & variational mean of the marginal $q(\vec x_t)$\\
$\vec S_t$ & variational covariance of the marginal $q(\vec x_t)$\\
\bottomrule
\end{tabular}
\end{table}

\subsection{Model joint distribution}
Here we define the joint distribution of the Gaussian processes $f$, the latent states $\vec x$ and the data $\vec y$.

The Gaussian processes have prior mean $\eta(\cdot)$ and prior covariances $k(\cdot,\,\cdot)$:
\begin{align}
	p(f_d(\cdot)) = \mathcal {GP} \big(\eta_d(\cdot),\, k_d(\cdot, \cdot)\big)\,\quad d=1\ldots D\,.
\end{align}
We note that placing a measure $p$ on the function $f$ causes some measure-theoretic discrepancies. Nonetheless, the derivation holds following a more theoretical consideration of the problem \citep{matthews2016sparse}, and the intuition given by our derivation is correct. 

The initial state is assumed to be drawn from a standard normal distribution
\begin{align}
p(\vec x_0) = \mathcal N(\vec 0,\,\vec I_D)\,.
\end{align}

The state transition depends on the Gaussian processes:
\begin{align}
	p(\vec x_t\given \vec x_{t-1}, \vec f(\cdot)) = \mathcal N \big(\vec x_t \given \vec f(\tilde{\vec x}_{t-1}),\, \sigma_f^2 \vec I_D\big)\,,
\end{align}

We assume a linear-Gaussian observation model:
\begin{align}
	p(\vec y_t \given \vec x_t) = \mathcal N\big(\vec y_t\given \vec W_{\!\!g} \vec x_t + \vec b_g, \sigma_g^2 \vec I_O\big)
\end{align}

The joint density is then
\begin{align}
p(\vec f, \vec X, \vec Y) = \prod_{d=1}^D p(f_d(\cdot)) \, p(\vec x_0) \,\prod_{t=1}^Tp(\vec y_t \given \vec x_t)\,\prod_{t=1}^Tp(\vec x_t \given \vec f, \vec x_{t-1})
\end{align}

\subsection{Approximate posterior distribution}
We will use variational Bayes to approximate the posterior distribution over $\vec f$ and $\vec X$, whilst simultaneously obtaining a bound on the marginal likelihood (the ELBO) which will be used to train the parameters of the model, including covariance function parameters, noise variances and the parameters $\vec W_{\!\!g}, \vec b_g$ of the linear output mapping. 

The posterior over Gaussian processes takes the form of a sparse GP. We introduce a series of $M$ variational inducing points $\vec Z = [\vec z_m]_{m=1}^M$ which lie in the same domain as $\tilde{\vec x}$. Following convention, the values of the $d^\textrm{th}$ function at those points are denoted $\vec u_d = [f_d(\vec z_m)]_{m=1}^M$, while evaluations from the prior as $\vec \eta_d = [\eta_d(\vec z_m)]_{m=1}^M$. Note that the variables $\vec u$ are not {\em auxiliary} variables, but part of the original model specification, being part of the GP. We assume a variational posterior of the form 
\begin{align}
	q(\vec U) = \prod_{d=1}^D \mathcal N\big(\vec u_d \given \vec \mu_d, \,\vec \Sigma_d\big)\,.
\end{align}
The remainder of the GPs conditioned on $\vec u$ are assumed to take the same form as the GP prior conditional. That is
\begin{align}
	q(f_d(\cdot) \given \vec u_d) = 
	p(f_d(\cdot) \given \vec u_d) = 
	\mathcal GP\big(
	\eta_d(\cdot) + k(\cdot, \vec Z)\vec K_{zz}^{-1}(\vec u_d - \vec \eta_d),\,
        k(\cdot, \cdot) - k(\cdot, \vec Z)\vec
	K_{zz}^{-1} k(\vec Z, \cdot)\big)\,.
\end{align} 
Marginalising with respect to $\vec u_d$ leads to our approximation to the GP: 
\begin{align}
	q(f_d(\cdot)) = \mathcal {GP}\big(\mu_d(\cdot), v_d(\cdot, \cdot)\big)\,,
\end{align}
with
\begin{align}
	\mu_d(\cdot) &= \eta_d(\cdot) + k(\cdot, \vec Z)\vec K_{zz}^{-1}(\vec \mu_d - \vec \eta_d)\,,\\
	v_d(\cdot, \cdot) &= k(\cdot, \cdot) - k(\cdot, \vec Z)\vec
        K_{zz}^{-1}[\vec K_{zz} - \vec \Sigma_d]\vec K_{zz}^{-1} k(\vec
        Z, \cdot)\,.
\end{align}

The approximation to the posterior over state trajectories is given a Gauss-Markov structure of the form
\begin{align}
	q(\vec X) = q(\vec x_0)\prod_{t=1}^Tq(\vec x_t\given \vec x_{t-1})\,,
\end{align}
where
\begin{align}
	q(\vec x_0) &= \mathcal N(\vec x_0\given \vec m_0,\,\vec L_0\vec L_0^\top)\\
	q(\vec x_t\given\vec x_{t-1}) &= \mathcal N(\vec x_t\given \vec A_t \vec x_{t-1},\,\vec L_t\vec L_t^\top)\,.
\end{align}

The complete set of variational parameters is then $\vec Z, \{\vec \mu_d, \vec \Sigma_d\}_{d=1}^D, \vec m_0, \vec L_0, \{\vec A_t, \vec L_t\}_{t=1}^T$. The parameters of $q(\vec X)$ are reconfigured to be the output of an RNN recognition model (see main text), whilst we optimise the parameters controlling $\vec f(\cdot)$ directly.

The joint posterior then factors as
\begin{align}
	q(\vec f(\cdot), \vec X) = \prod_{d=1}^Dq(f_d(\cdot)) q(\vec X)\,.
\end{align}

\subsection{The ELBO}
Having specified the forms of the model and the approximate posterior, we are ready to derive the ELBO. Following the standard variational Bayes methods, we write
\begin{align}
	\textsc{ELBO} = \mathbb E_{q(\vec X)q(\vec f(\cdot))} \left[\log \frac{p(\vec Y\given \vec X)p(\vec X\given \vec f(\cdot))}{q(\vec X)} \frac{p(\vec f(\cdot))}{q(\vec f(\cdot))}\right]\,.
\end{align}
We will split the ELBO into four parts, dealing with each in turn:
\begin{align}
	\nonumber
	\textsc{ELBO} = &\underbrace{\mathbb E_{q(\vec X)} \big[\log p(\vec Y\given \vec X)\big]}_\text{part 1}
        + \underbrace{\mathbb E_{q(\vec X)q(\vec f(\cdot))} \big[\log p(\vec X\given \vec f(\cdot)) \big]}_\text{part 2}\\
	&- \underbrace{\mathbb E_{q(\vec X)} \big[\log q(\vec X)\big]}_\text{part 3}
	+ \underbrace{\mathbb E_{q(\vec f(\cdot))} \big[\log \frac{p(\vec f(\cdot))}{q(\vec f(\cdot))}\big]}_\text{part 4}\,.
\end{align}

\paragraph{Part 1} This expression can be computed straight-forwardly in closed form due to our choice of a linear-Gaussian emission $g(\vec x)$.  Let $\vec m_t, \vec \Sigma_t$ be the marginals of $q(\vec x_t)$ computed via the recursion, and recall the form of the linear emission function $g(\vec x_t) = \vec W_{\!\!g} \vec x_t + \vec b_g$
\begin{align}
	\nonumber
	\mathbb E_{q(\vec X)} \big[\log p(\vec Y\given \vec X)\big] &= 
	\mathbb E_{q(\vec X)} \Big[\sum_{t=1}^T \log \mathcal N\big(\vec y_t \given g(\vec x_t), \sigma^2_g\big)\Big]\\
	\nonumber
	&=\sum_{t=1}^T\mathbb E_{q(\vec x_t)} \Big[ \log \mathcal N\big(\vec y_t \given \vec W_{\!\!g} \vec x_t + \vec b_g, \sigma^2_g\big)\Big]\\
	&=\sum_{t=1}^T \log \mathcal N\big(\vec y_t \given \vec W_{\!\!g} \vec m_t + \vec b_g, \sigma^2_g\big) - \tfrac{1}{2\sigma^2_n}\textrm{tr}(\vec W_{\!\!g}^\top\vec W_{\!\!g}\vec\Sigma_t)\,.
\end{align}
In practise we defer this simple computation to the \verb|variational_expectations| functionality in GPflow \citep{matthews2017gpflow}. 
\paragraph{Part 2} This expression cannot be computed in closed form without restriction to the RBF kernel as in \citep{frigola2015bayesian}. We eliminate the integral with respect to $\vec f$ here, and then use the reparameterisation trick to estimate the integral with respect to $\vec X$ (see main text). 
\begin{align}
	\nonumber
	\text{part 2} &= 
	\mathbb E_{q(\vec X)q(\vec f(\cdot))} \big[\log p(\vec X\given \vec f(\cdot)) \big]\\
	\nonumber
	&= \mathbb E_{q(\vec X)q(\vec f(\cdot))} \big[\log p(\vec x_0)\prod_{t=1}^T\mathcal N\big(\vec x_t\given \vec f(\tilde{\vec  x}_{t-1}),\,\sigma_f^2 \vec I_D\big) \big]\\
	\nonumber
        &=\mathbb E_{q(\vec x_0)}\big[\log p(\vec x_0)\big] + \mathbb E_{q(\vec X)q(\vec f(\cdot))} \Big[\sum_{t=1}^T\sum_{d=1}^D\log \mathcal N\big(\vec x_t^{(d)}\given f_d(\tilde{\vec  x}_{t-1}),\,\sigma_f^2 \big) \Big]\\
&=\mathbb E_{q(\vec x_0)}\big[\log p(\vec x_0)\big] + \mathbb E_{q(\vec X)} \Big[\sum_{t=1}^T\sum_{d=1}^D\log \mathcal N\big(\vec x_t^{(d)}\given \mu_d(\tilde{\vec  x}_{t-1}),\,\sigma_f^2 \big) - \tfrac{1}{2}\sigma_f^{-2}v_d(\tilde{\vec  x}_{t-1}, \tilde{\vec  x}_{t-1})\Big],
\end{align}
which matches the term in the main text. 

\paragraph{Part 3} This corresponds to the entropy of $q(\vec X)$. It is straightforward to derive:
\begin{align}
	-\mathbb E_{q(\vec X)} \big[\log q(\vec X)\big]  = \textsc{H}[q(\vec X)] = \frac{(T+1)D}{2}\log(2\pi e) +\sum_{t=0}^T\log(\det(\vec L_t))\,.
\end{align}

\paragraph{Part 4} This final part is the Kullback-Leibler divergence between the prior and (approximate) posterior GPs. We first note that it can be written as a sum across dimensions $d$, and then that each GP $f_d(\cdot)$ can be factored into two parts: $p(f_d(\cdot)\given \vec u_d) p(\vec u_d)$ and similarly for $q$. This results in
\begin{align}
	\nonumber
	\mathbb E_{q(\vec f(\cdot))} \left[\log \frac{p(\vec f(\cdot))}{q(\vec f(\cdot))}\right] &= 
	\sum_{d=1}^D \mathbb E_{q(f_d(\cdot))} \left[\log \frac{p(f_d(\cdot))}{q(f_d(\cdot))}\right]\\ &= 
	\sum_{d=1}^D \mathbb E_{q(f_d(\cdot)\given \vec u_d)q(\vec u_d)} \left[\log \frac{p(f_d(\cdot)\given \vec u_d)p(\vec u_d)}{q(f_d(\cdot)\given \vec u_d)q(\vec u_d)}\right]\,.
\end{align}
Since we have defined the posterior conditional $q(f_d(\cdot)\given\vec u_d$) to match the prior conditional, the two terms cancel, resulting in
\begin{align}
	\nonumber
	\mathbb E_{q(\vec f(\cdot))} \left[\log \frac{p(\vec f(\cdot))}{q(\vec f(\cdot))}\right] &= 
	\sum_{d=1}^D \mathbb E_{q(\vec u_d)} \left[\log \frac{p(\vec u_d)}{q(\vec u_d)}\right] \\
	&=-\sum_{d=1}^D \textsc{KL}\big[q(\vec u_d) || p(\vec u_d)\big]\,.
\end{align}

Since the result is a Kullback-Leibler divergence between two finite-dimensional normal distributions, it is computed straightforwardly. 

Although this notation is somewhat sloppy (since the sets of variables $f_d(\cdot)$ and $\vec u_d$ overlap), the result is correct. \citet{matthews2017thesis} contains a more careful and significantly more technical derivation.

\clearpage
\section{Full visualisation of synthetic 1D dataset}
\label{sec:full_1d_viz}
\begin{figure}[!h]
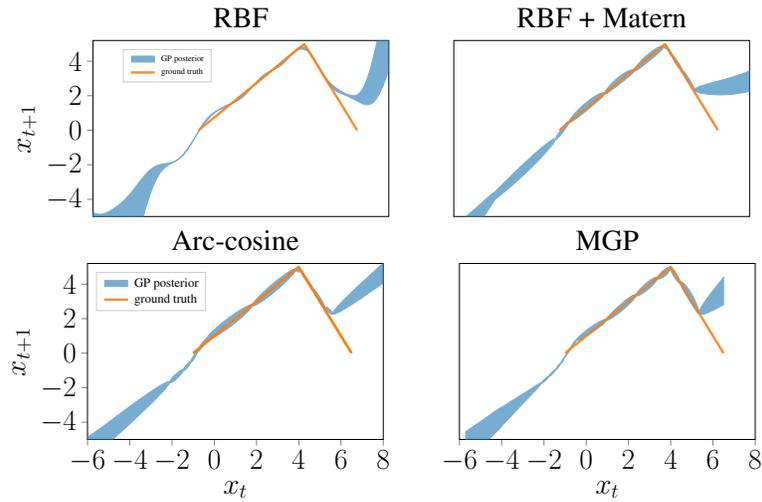

		\centering
	\begin{tabular}{cc}

	\scalebox{.53}{
		\input{\figpath/one_d/full_RBF_0.1_0.01_transition_model.tex}
	} &
	\scalebox{.53}{
		\input{\figpath/one_d/full_RBFMatern_0.1_0.01_transition_model.tex}
	}\\
	\scalebox{.53}{
		\input{\figpath/one_d/full_ArcCosine0_0.1_0.01_transition_model.tex}
	} &
	\scalebox{.53}{
		\input{\figpath/one_d/full_MGP_0.1_0.01_transition_model.tex}
	}
	\end{tabular}
	\caption{Visualisation of the learned GP transition functions across a greater domain of the function. It can be seen that all models revert to the mean function (defined as the identity function) away from the data. The short lengthscales of the Arc-cosine and MGP (compounded with a Matern kernel) that are used to fit the kink of the true transition function mean that they almost instantaneously revert to the mean function. The longer length scales of the RBF-containing kernels mean that we revert much more slowly to the mean.}
	\label{fig:one_d_full}
\end{figure}

\clearpage

\section{Learnt latent states for cart-pole}
Below we provide the learnt latent states for the cart-pole dataset with observed and hidden velocities. It is worth noting that the model has recovered similar structure for both cases.

\begin{figure}[!h]
\scalebox{.52}{
\input{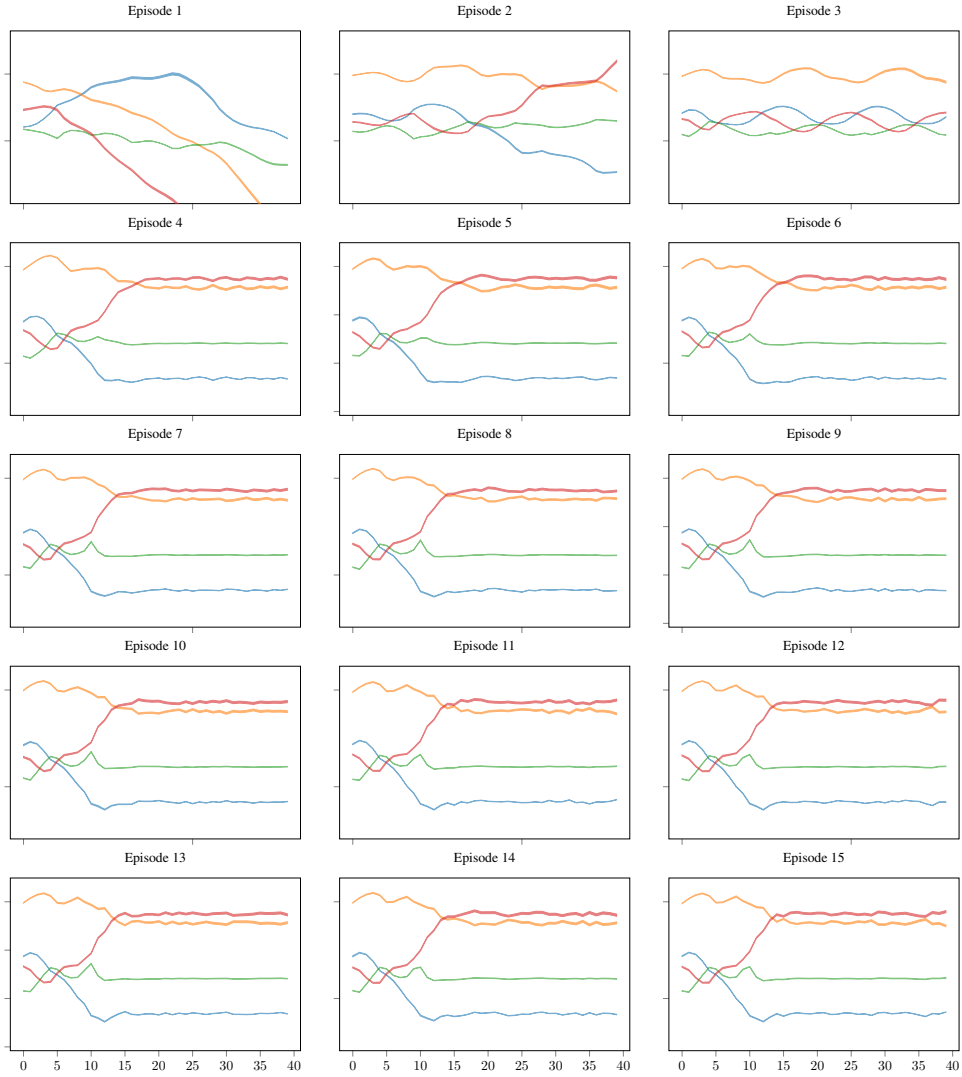}
}
\caption{Learnt latent states for the cart-pole dataset with observed velocities.}
\end{figure}

\clearpage

\begin{figure}[!h]
\scalebox{.52}{
\input{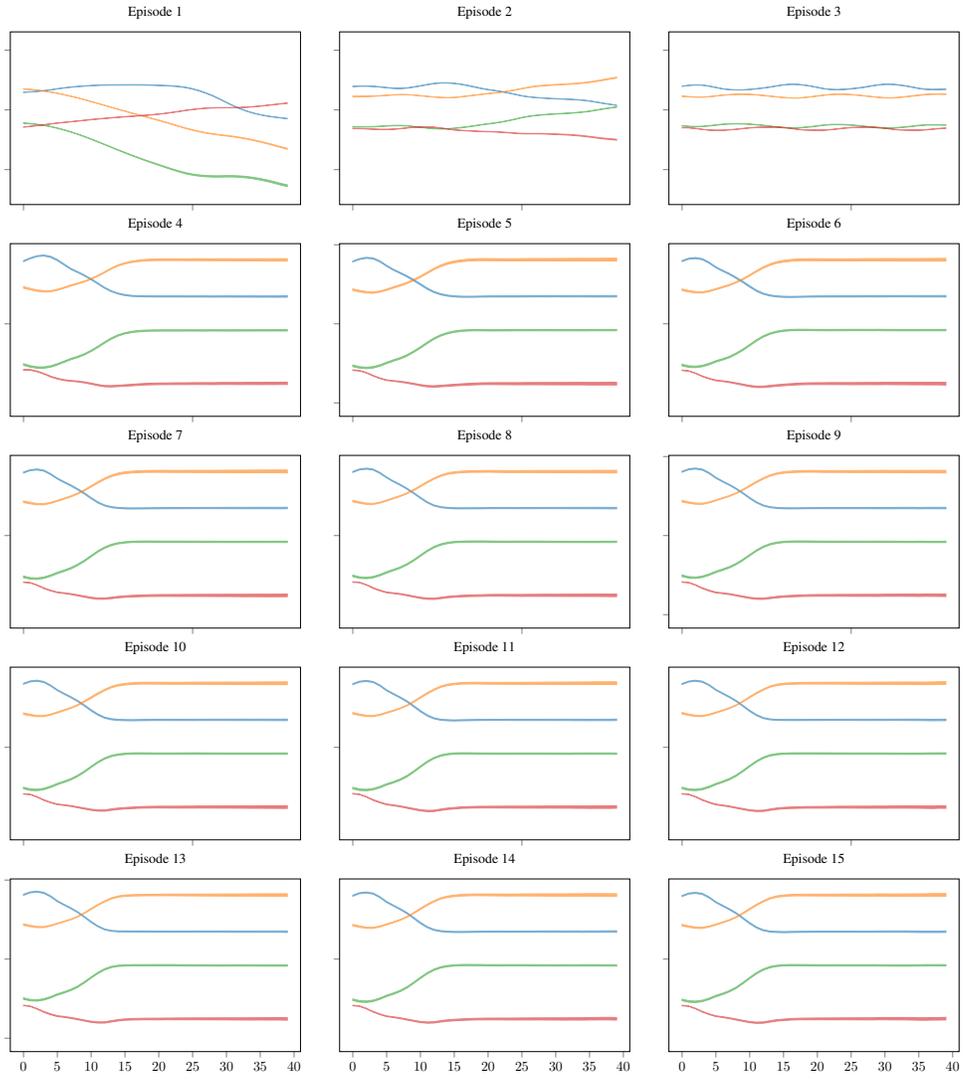}
}
\caption{Learnt latent states for the cart-pole dataset with hidden velocities.}
\end{figure}

\clearpage
\section{Cart-pole training data fitting}\label{append:fittings}

Below we provide detailed fittings on the training episodes for the cart-pole dataset.
\begin{figure}[!h]
\scalebox{.52}{
\input{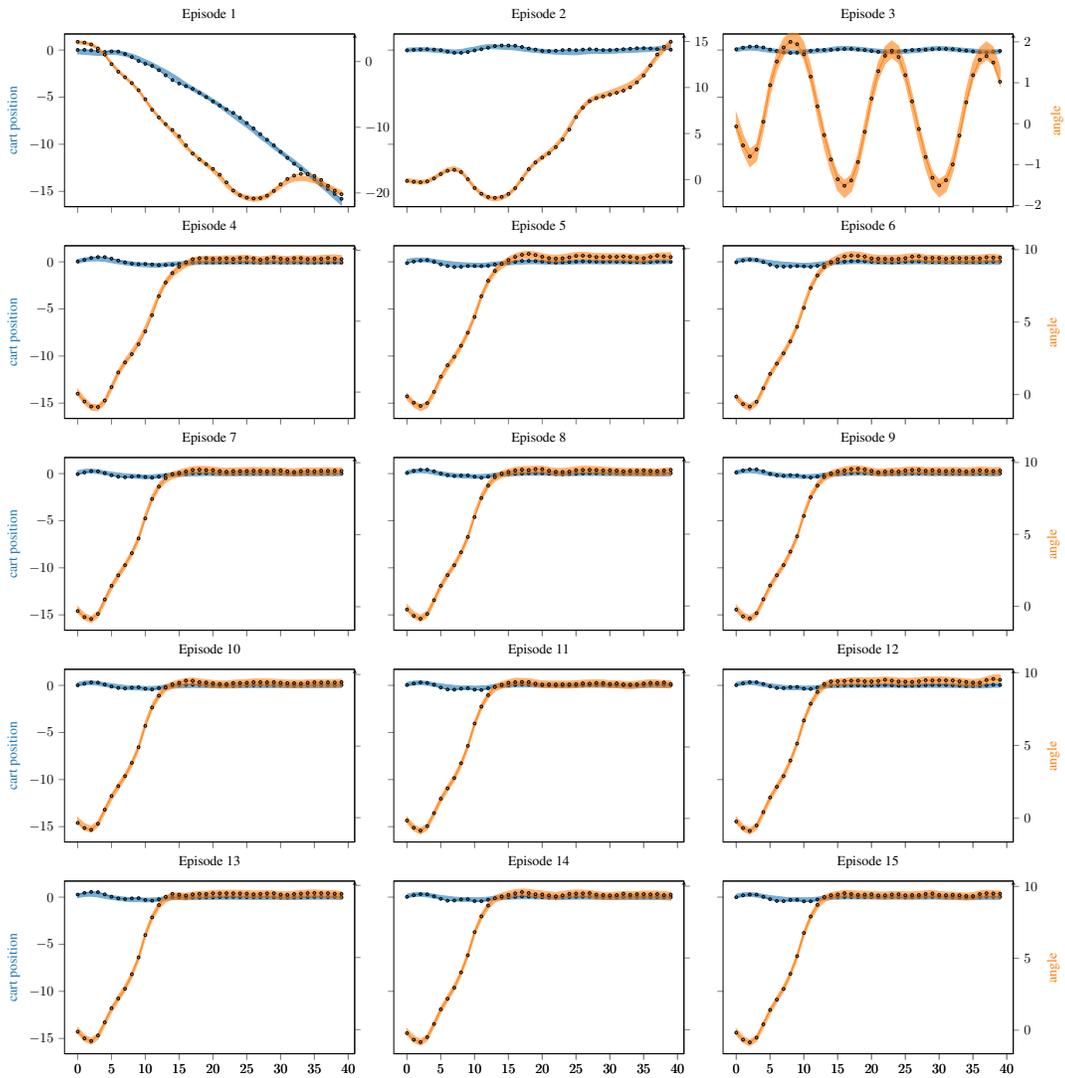}
}
\caption{Detailed fittings per episode for the cart-pole dataset with observed velocities.}
\end{figure}

\begin{figure}
\scalebox{.52}{
\input{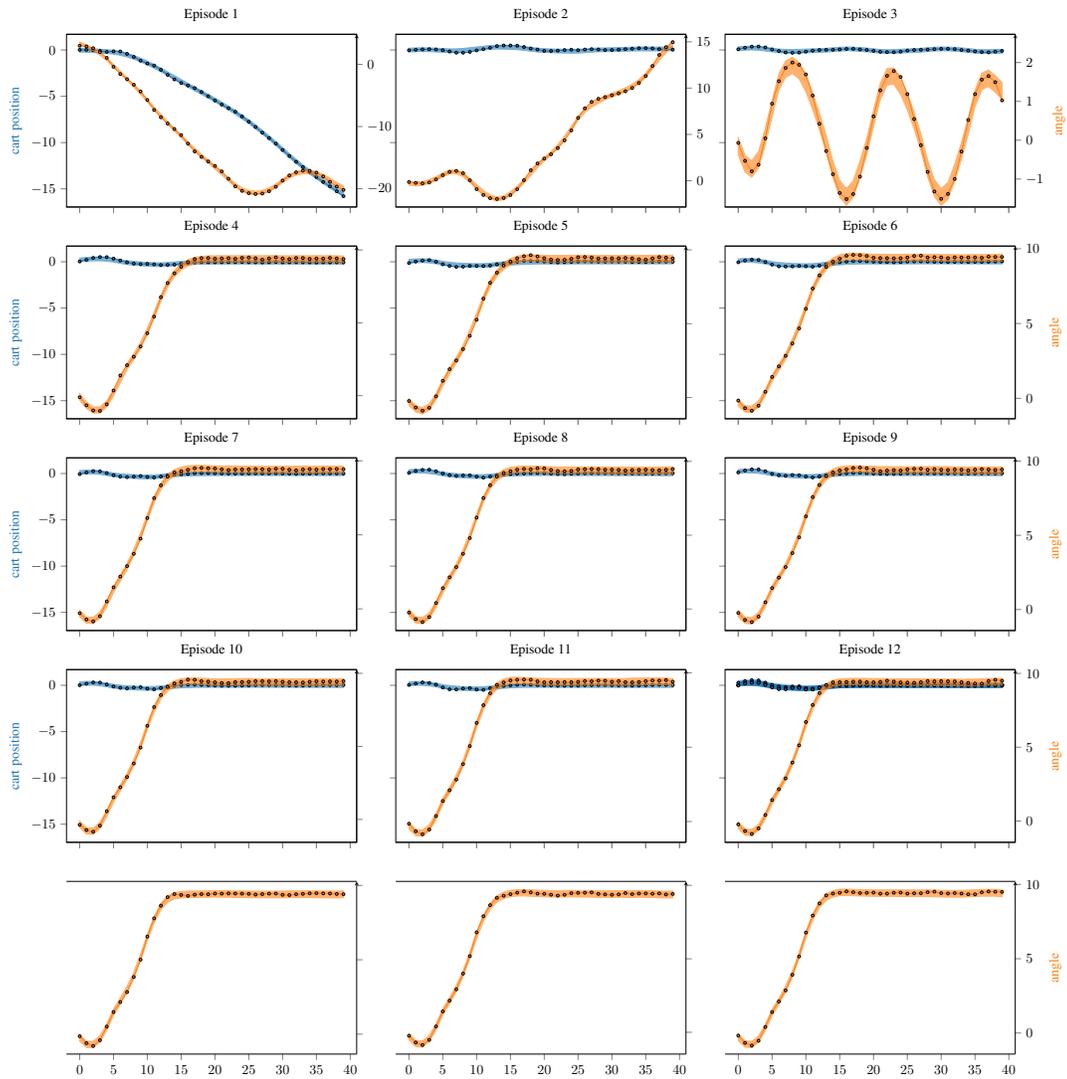}
}
\caption{Detailed fittings per episode for the cart-pole dataset with hidden velocities.}
\end{figure}


\end{appendices}
\end{document}